Original Paper

# An Urban Population Health Observatory for Disease Causal Pathway Analysis and Decision Support: Underlying Explainable Artificial Intelligence Model

Whitney S Brakefield[1,2], MSc, PhD; Nariman Ammar[1,3], MSc, PhD; Arash Shaban-Nejad[1], MPH, PhD

[1]Center for Biomedical Informatics, Department of Pediatrics, College of Medicine, University of Tennessee Health Science Center, Memphis, TN, United States

[2]Bredesen Center for Data Science, University of Tennessee, Knoxville, TN, United States

[3]Ochsner Xavier Institute for Health Equity and Research, Ochsner Clinic Foundation, New Orleans, LA, United States

**Corresponding Author:**
Arash Shaban-Nejad, MPH, PhD
Center for Biomedical Informatics
Department of Pediatrics, College of Medicine
University of Tennessee Health Science Center
50 N Dunlap Street, R492
Memphis, TN, 38103
United States
Phone: 1 901 287 5863
Email: ashabann@uthsc.edu

## Abstract

**Background:** Many researchers have aimed to develop chronic health surveillance systems to assist in public health decision-making. Several digital health solutions created lack the ability to explain their decisions and actions to human users.

**Objective:** This study sought to (1) expand our existing Urban Population Health Observatory (UPHO) system by incorporating a semantics layer; (2) cohesively employ machine learning and semantic/logical inference to provide measurable evidence and detect pathways leading to undesirable health outcomes; (3) provide clinical use case scenarios and design case studies to identify socioenvironmental determinants of health associated with the prevalence of obesity, and (4) design a dashboard that demonstrates the use of UPHO in the context of obesity surveillance using the provided scenarios.

**Methods:** The system design includes a knowledge graph generation component that provides contextual knowledge from relevant domains of interest. This system leverages semantics using concepts, properties, and axioms from existing ontologies. In addition, we used the publicly available US Centers for Disease Control and Prevention 500 Cities data set to perform multivariate analysis. A cohesive approach that employs machine learning and semantic/logical inference reveals pathways leading to diseases.

**Results:** In this study, we present 2 clinical case scenarios and a proof-of-concept prototype design of a dashboard that provides warnings, recommendations, and explanations and demonstrates the use of UPHO in the context of obesity surveillance, treatment, and prevention. While exploring the case scenarios using a support vector regression machine learning model, we found that poverty, lack of physical activity, education, and unemployment were the most important predictive variables that contribute to obesity in Memphis, TN.

**Conclusions:** The application of UPHO could help reduce health disparities and improve urban population health. The expanded UPHO feature incorporates an additional level of interpretable knowledge to enhance physicians, researchers, and health officials' informed decision-making at both patient and community levels.

**International Registered Report Identifier (IRRID):** RR2-10.2196/28269

*(JMIR Form Res 2022;6(7):e36055)* doi: 10.2196/36055

XSL·FO
**RenderX**

## Introduction

### Background

Enhanced health surveillance systems for chronic disease support could mitigate factors that contribute to the incline of morbidity and mortality of diseases such as obesity. Obesity is linked to increased overall mortality and has reached pandemic proportions, being responsible for approximately 2.8 million deaths annually [1,2]. Obesity represents an excessive and abnormal accumulation of body fat, which leads to adverse health effects that impose a health and financial toll on individuals and society [2]. More than half of the US population has at least one chronic condition, and 27% are living with multimorbidity [3]. These conditions cause more than 1.7 million deaths per year in the United States, where obesity is associated with the top leading causes of death (eg, diabetes, heart disease, stroke, and cancer) [4].

Neighborhood factors such as socioenvironmental determinants of health (SDoH) significantly contribute to these statistics [5-8]. Implementation of an intelligent health surveillance platform that incorporates SDoH can improve preparedness, prevention, and management of this obesity pandemic by assisting in the implementation of effective treatment and interventions.

Health surveillance involves the "ongoing systematic collection, analysis, and interpretation of data essential to the planning, implementation, and evaluation of public health practice, closely integrated with the timely dissemination of these data to those who need to know" [9]. Researchers have aimed to develop chronic health surveillance systems to assist in chronic health decision-making [10-16]. The World Health Organization (WHO) developed a conceptual framework for an Urban Public Health Observatory (UPHO) comprised of 3 domains: mission, governance, and knowledge and intelligence, the latter of which incorporates a data management component [17]. This framework from WHO therefore provides a strategic model for health surveillance.

Many current digital health solutions and electronic health record (EHR) systems lack the ability to incorporate machine learning algorithms into their decision-making process, and even if they do, the algorithms used do not have appropriate capabilities to explain the suggested decisions and actions to human users [18]. Machine learning approaches, so-called black-box statistics, should be trustworthy, transparent, interpretable, and explainable when making decisions in the clinical or health science setting [18-20]. A system's explanation constitutes its interpretability [18,20-22]. Explainable AI (XAI) increases the intelligence delivered to the user by providing explanations, thereby enhancing the interpretability of outcomes and findings. Researcher efforts have been shifting toward applying algorithms that can aid in explaining the results of machine learning models. For instance, the SHAP (Shapley Additive Explanations) analysis [23] is an approach that assigns each model feature an importance score for making a particular prediction. Compared to traditional feature importance analyses, the novelty of SHAP lies in its ability to assess importance at the individual patient level. In this paper, we propose a novel approach to explainability that uses knowledge graphs as a semantic infrastructure explainable by design and enriches those graphs with results from machine learning algorithms as metrics and scores. The semantic causal relationships on the graph provide contextual knowledge around a population, and the metrics support those relationships, which provides 2 levels of evidence: knowledge level and statistical level.

We implement a UPHO platform as a knowledge-based surveillance system that provides better insight to improve decision-making by incorporating SDoH and providing XAI and interpretability functions [24]. Our UPHO consists of 3 layers: data, analytics, and application. In this work, we refine the initial design by incorporating data management, knowledge, and intelligence domains (Figure 1) that are in alignment with the conceptual model by WHO and a focus on the semantics layer.

**Figure 1.** Expanded Urban Population Health Observatory framework. CDC: US Centers for Control and Prevention; USDA: US Department of Agriculture; KG: knowledge graph; UPHO: Urban Population Health Observatory.

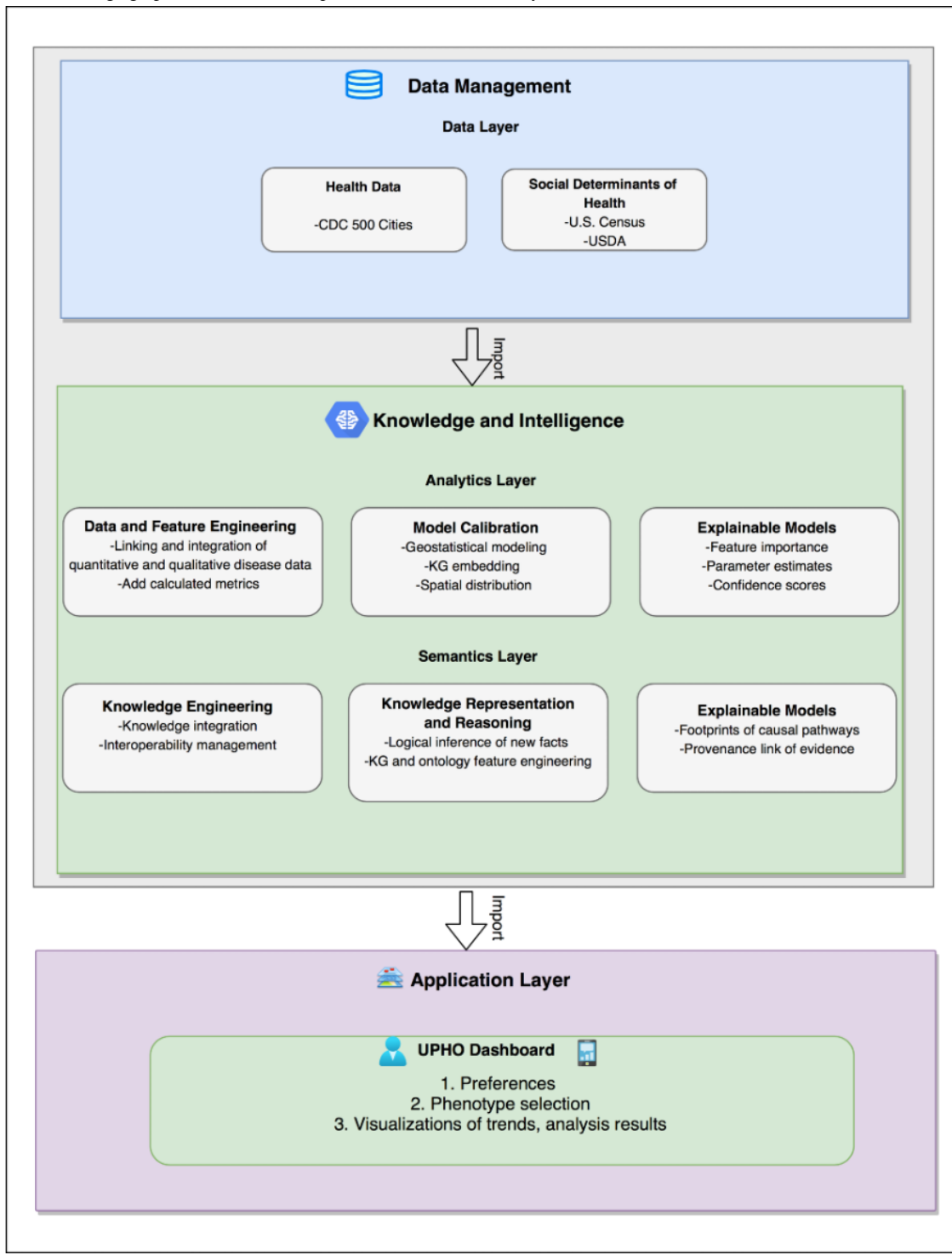

## Objectives

The objectives of this article are to (1) expand UPHO by incorporating a semantics layer, (2) cohesively employ machine learning and semantic/logical inference to provide measurable evidence and detect pathways that lead to undesirable health outcomes, (3) provide clinical case scenarios and design case studies on identifying SDoH associated with obesity prevalence, and (4) provide a dashboard design that demonstrates the use of UPHO in the context of obesity, using the provided case scenario.

## *Methods*

### UPHO Expansion

Figure 1 shows the expansion of the UPHO to incorporate the semantics layer. In the following section, we provide a detailed description of the UPHO platform expansion design.

### Data Management Domain

The data management domain comprises the data layer. The UPHO collects population-level health and SDoH data and individual-level clinical and demographic data from EHRs through regional registries.

## Data Layer

To obtain population-level health data, we used the US Centers for Control and Prevention (CDC) 500 Cities Behavior Risk Factors Surveillance System, which includes data regarding chronic diseases and their behavioral risk factors [25]. These variables are model-based estimates of crude prevalence among adults aged ≥18 years in 2018. We extracted variables pertaining to obesity, lack of physical activity, lack of insurance, and diabetes mellitus at the census tract level.

We extracted population-level SDoH variables that pertain to food insecurity, transportation, and socioeconomic stability at zip code, census tract, census block, and census block group levels from the US Census Bureau 2018 American Community Survey [26] and the US Department of Agriculture Research Atlas [27].

## Knowledge and Intelligence Domain

### Analytics Layer

The analytics layer pulls raw data from different sources in the data layer and analyzes it to classify it, predict new relations, conduct spatial pattern detection, and calculate new metrics. The analytics layer also performs feature engineering by deriving new metrics and using them to enrich the original data sets.

### Semantics Layer

The stages of the UPHO semantics layer are shown in Figure 2. In the knowledge representation and axiomatization stages, we use semantic web technologies to develop several domain and application ontologies from relevant domains of interest to provide necessary contextual knowledge. Ontologies are systematic representations of knowledge that can be used to integrate and analyze large amounts of heterogeneous data, thereby allowing classification of knowledge [28]. In those ontologies, we define concept hierarchies and rule axioms by using existing domain knowledge, such as the WHO/CDC guidelines, and federal and local sources. We develop new ontologies by reusing several existing domain ontologies. For this study, we adopted concepts, properties, and axioms from 5 different ontologies that we used in our prior work [29-32], specifically (1) disease ontology (DO) [31] (eg, obesity, diabetes); (2) the Childhood Obesity Prevention (Knowledge) Enterprise (COPE) ontology [29] that defines SDoH concepts such as socioeconomic issues (eg, food deserts, income) and behavioral issues (eg, lack of physical activity, purchasing preference); (3) geographical information system ontology (GISO) [32] (eg, zip code, census tract); (4) health indicators ontology (HIO) [32]; and (5) the adverse childhood experiences (ACEs) ontology (ACESO), which defines concepts related to ACEs, health outcomes (eg, mental and physical health), interventions, and SDoH, including axioms that define issues like lack of transportation (eg, limited access to a vehicle and limited access to public transit) and food and how they affect routine follow-up activities (eg, missing medical appointments) [30].

We start our semantic analysis using concepts defined in our ontologies and web services to align concepts to actual data resources, allowing us to construct a population knowledge graph structure that abides by an ontology and contains both data and concepts [33]. We enrich that knowledge graph using a logical reasoner that uses facts derived from existing knowledge, new knowledge extracted from the analytics layer, and the generic rule axioms defined in the domain ontologies that trigger specific actions under certain conditions.

**Figure 2.** Urban Population Health Observatory semantic layer framework.

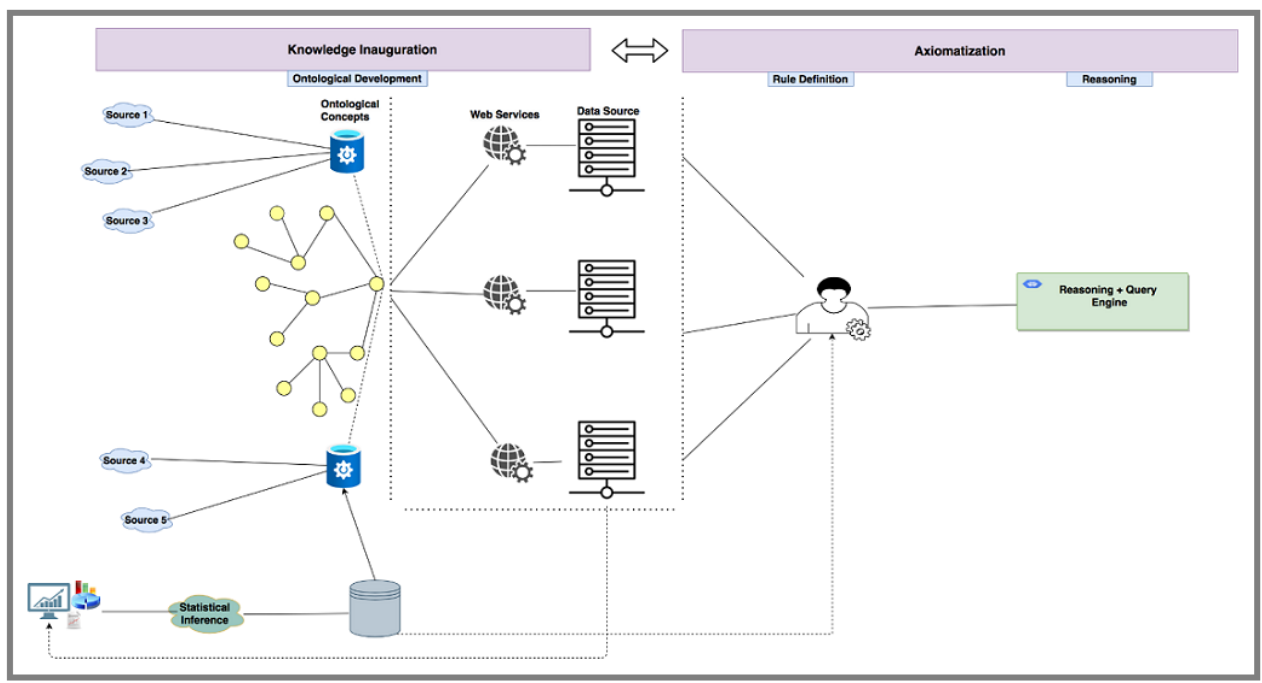

## Explainable AI

An effective explainable system accounts for the target user group (eg, physician, researcher). Knowledge of the end user is very important for the delivery of decisions, recommendations, and actions. Each analytics and semantics layer contains an explainability component that can be leveraged in the uppermost health applications layer. To maintain features such as data integration, XAI, and interpretability, we must achieve interoperability by using semantics and ontologies. Explanations adaptable to the user can decrease errors in interpretation by enhancing the interpretability of outcomes and findings.

## Application Domain

The UPHO platform can be used as a basis to develop several applications, some of which we have already developed, including dashboards [24], mobile health (mHealth) apps [34], digital assistants, and recommender systems [35]. In this article, we leverage the UPHO to implement a dashboard for real-time surveillance. By accessing dynamic knowledge discovered through the UPHO, the dashboard can provide real-time early warnings that are based both on content and context. The platform is accessible to policymakers, physicians, researchers, public health officials, and the public.

## Clinical Scenarios

The following sections present 2 clinical case scenarios that focus on a physician and a researcher as users to demonstrate the methodology used in the knowledge and intelligence domain layers and the corresponding dashboard design in the application.

Scenario 1: A physician seeks an effective intervention for an adult African American patient diagnosed with obesity. The physician focuses on how SDoH in the patient's neighborhood can influence the doctor's management plans.

Scenario 2: A researcher investigating the impact of SDoH on obesity seeks an effective intervention for the adult obese populations in Memphis, TN.

## Analytics Layer: Machine Learning Model Development

We trained a machine learning–based support vector regression (SVR) machine model [36,37]. We linked CDC 500 Cities population-level obesity and behavior data [25] to the population-level social DoH [26,27] data set at the census tract level in Memphis, TN. We analyzed 8 features and used a Spearman rank test to assess the positive or negative relationship between each feature. We used a variance inflation factor (VIF) to detect multicollinearity between features. To examine patient neighborhood-level exposure, we used SHAP analysis. Table 1 shows the summary statistics for features considered for this study. We trained our support vector model on the 85% of randomly selected training data and tested the model on the 15% of remaining data to ensure the generalizability of the model, and we applied the linear kernel function. We scaled our data to have a mean of zero and standard deviation of one. We applied the grid search optimization method to seek optimal hyperparameters to improve model performance using the Caret package in R software (R Foundation for Statistical Computing) [38]. In addition, to avoid overfitting 5-fold cross-validation was applied to the training data set. We used root mean square error (RMSE) and $R^2$ to evaluate the performance of the model.

**Table 1.** Summary statistics for obesity and related risk factors in Memphis, TN, census tract (n=178 census tracts).

| Features | Operationalization | Original, mean (SD) | Training, mean (SD) | Test, mean (SD) |
|---|---|---|---|---|
| Obesity | Model-based estimate for crude prevalence of obesity among adults aged ≥18 years, 2018 | 37.50 (7.84) | 37.42 (7.54) | 37.97 (6.95) |
| Low access to supermarket | Count of low income population more than half mile from a supermarket in the census tract | 1382.20 (108.37) | 1345.68 (967.83) | 1616.17 (1120.23) |
| Black | Percentage of population that is Black or African American | 63.17 (32.70) | 62.22 (33.04) | 63.72 (31.88) |
| Poverty | Percentage of population living below the federal poverty line | 28.65 (16.28) | 28.27 (16.18) | 31.06 (17.06) |
| Unemployment | Percentage of unemployed population | 15.73 (9.31) | 15.97 (9.67) | 14.16 (6.52) |
| High school diploma | Percentage of population 25 years or older without high school diploma | 10.38 (6.59) | 10.23 (6.70) | 11.35 (5.89) |
| Lack of physical activity | Model-based estimate for crude prevalence of lack of physical activity among adults aged ≥18 years, 2018 | 36.16 (9.80) | 35.97 (9.79) | 37.34 (9.99) |
| Crime | Crime rate per thousand people | 350.20 (126.26) | 160.99 (337.65) | 111.93 (80.40) |
| Lack of access to insurance | Model-based estimate for crude prevalence of lack of insurance among adults aged ≥18 years, 2018 | 20.21 (6.78) | 20.10 (6.81) | 20.88 (6.67) |

## Semantics Layer: Knowledge Graph Generation

We followed the following ordered steps to generate the semantics layer knowledge graph from concepts defined in our domain ontologies.

1. We use concepts, relations, and axioms from domain ontologies to construct a preliminary population knowledge graph. For our scenario, we start by adding a dummy node that represents either a patient or a population ([Figure 3](#)). We begin connecting that node to concepts like disease, risk factors, physical characteristics, etc. For example, obesity falls into the disease type represented by the *isA* relation in [Figure 3](#), where *isA* reflects a subtype. For instance, SDoH *isA RiskFactr*, and lackOfTransporation is an SDoH subtype. These different hierarchies are encoded in the ACESO ontology. We also add a prefix before each type to reflect the namespace where that concept is defined (eg, the term DO:Disease reflects that the concept disease is defined in the DO). Relations can also reflect the properties of a node. For example, a patient *livesIn* 38127, which *isA* zip code, and that zip code has 8 census tracts of the type *CensusTract*, defined in the GISO ontology.

2. We populate the generated graph structure with evidence from the data layer. For instance, our data set contains a variable that shows the prevalence of obesity as a percentage metric in specific neighborhoods. We use that information to add edges to our graph that link obesity (as a disease) to prevalence (as a metric).

3. We further enrich and refine the initial graph by performing knowledge engineering using the logical reasoner ([Figure 2](#)) and feature engineering using the results from the analytics layer. The logical reasoner uses a set of rule axioms to perform logical inference on concepts already existing in the graph. For instance, our COPE ontology encodes epidemiological causal axioms that link SDoH to negative health outcomes. [Textbox 1](#) shows how we encode the generic axioms R1-R3. When we combine those generic rule axioms with facts about a specific census tract, we can infer all the risk factors associated with living in that census tract, eg, knowing the facts F1 can tell us that the population living in that area may be exposed to risk factors that lead to obesity.

4. After performing the logical inference on the initial graph structure, we incorporate new nodes and edges in the graph corresponding to new concepts (eg, the lackOfPhysicalActivity concept from the COPE ontology) or new relations (eg, *isExposedTo*). The knowledge graph refinement is an iterative process, so we can repeat step 2 until we reach a stable state of the graph after which we can populate the graph with more evidence from our engineered data that we pull from the analytics layer. For that purpose, we use the population-level data about SDoH risk factors collected from US Census, CDC, and USDA. For instance, to capture the lack of physical activity we use the CDC 500 Cities data set. The machine learning analysis performed by the analytics layer provides edges that pertain to prediction (eg, *isPredictorOf*, [Figure 3](#)). The final knowledge graph is shown in [Figure 3](#), which provides a generic view of all possible assumptions we can make about this patient or population.

5. To gather the most important information from this graph, a user can trace a specific pathway based on both logical inference and machine learning results. The red arrows in [Figure 3](#) reflect the pathway in our scenario.

**Figure 3.** Knowledge graph that links concepts defined in domain ontologies (eg, GISO: CensusTract) to data resources stored in databases (eg, percentage Black population) or those derived from the analytics layer. The upper part of the figure shows the nodes and edges produced through semantic inference during the knowledge engineering phase. The lower part of the figure shows the nodes and edges added through ML analysis during the feature engineering phase. GISO: geographical information system ontology; HIO: health indicators ontology; ACESO: adverse childhood experiences ontology; COPE: Childhood Obesity Prevention (Knowledge) Enterprise; DO: disease ontology.

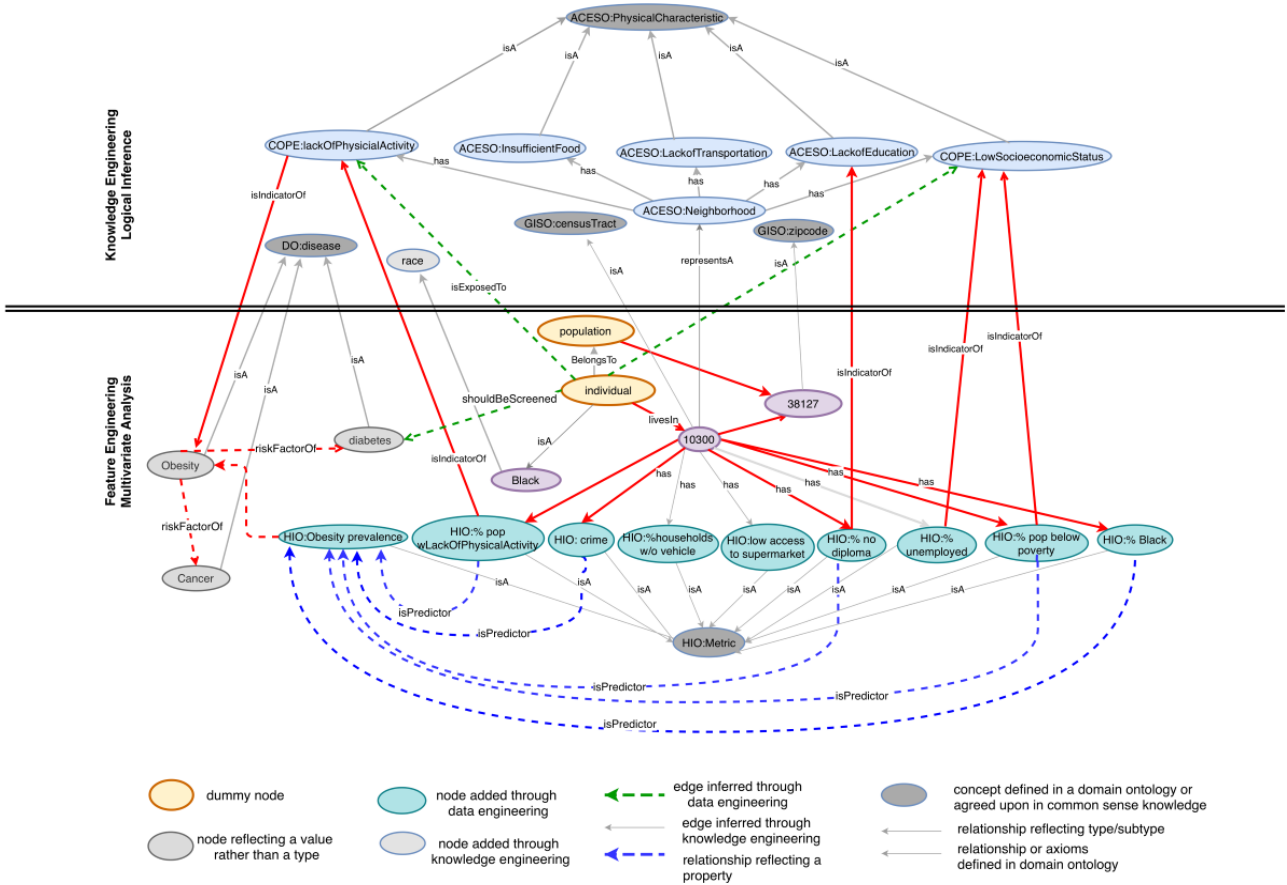

**Textbox 1.** Encoding axioms as general rules, initial facts, or new facts derived from feature engineering or logical reasoning.

Generic rule axioms

- COPE:lackOfPhysicialActivity *leadsTo* DO:Obesity (R1)
- %ObesityPrevalence:Metric *isHealthIndicatorFor* DO:Obesity (R2)
- Obesity:Disease *isRiskFactorOf* Diabetes:Disease (R3)

Facts

- individual:Patient *livesIn* "10300":CensusTract (F1)
- "10300":CensusTract *has* "49":%PopWLackOfPhysicialActivity (F2)
- "10300":CensusTract *has* "21":%PopNoHighSchoolDiploma (F3)
- "10300":CensusTract *has* "60":%UnderPovertyLine (F4)
- "10300":CensusTract *has* "46":%ObesityPrevalence (F5)

Feature engineered through multivariate analysis

- %PopWLackOfPhysicialActivity:Metric *isPredictorOf* ObesityPrevalence:Metric [using F2-F5] (F6)

Logical reasoning

- individual:Patient *isExposedTo* LackPhysicalActivity:PhysicalCharacteristic [using F1 and F2] (F7)
- individual:Patient *shouldBeScreenedFor* Diabetes:Disease [using R1, R2, R3, F6, F7]

### Ethics Approval

No ethics review board assessment was required for this study because we used publicly available data.

## Results

### Machine Learning Analysis

The significant Spearman rank coefficient and VIF of the 7 features included in this study are shown in Table 2. Any feature exhibiting a VIF of greater than 10 was removed. For the SVR model results, we obtained an RMSE of 0.312 for the training set and 0.203 for the test set, while the $R^2$ for the training set was 0.91 and that for the testing data set was 0.95 . Since the model provides similar results for training and test data sets, the proposed model does not overfit. The SVR feature importance results range on a scale of 0 to 100, and the greater the score, the most important the feature (Table 3). We found that the percentage of the population lacking physical activity, percentage of population below poverty level, percentage of population without high school diploma, percentage of population unemployed, and percentage of Black population were the most important variables when predicting obesity prevalence in Memphis, TN.

Figure 4 shows SHAP's value plot of feature contribution at the patient neighborhood level (census tract: 10300), which indicates the most important features such as the percentage of the population that lack physical activity and the percentage of population below the poverty level, from the point of view of the prediction of obesity prevalence in the patient neighborhood. The lack of physical activity and poverty had the largest positive (increased) contributions to obesity prevalence. On the other hand, the population of low income and more than a half-mile from the supermarket showed a negative (decreased) contribution but was the least important variable when predicting the patient neighborhood obesity prevalence. The knowledge extracted from our analysis will be used to detect the obesity prevalence pathways, which are defined by the top 5 most important features.

**Table 2.** Spearman rank coefficient and variance inflation factor for each feature.

| Features | Spearman rank coefficient | VIF[a] |
|---|---|---|
| Low access to supermarket | 0.37 | 1.70 |
| Black | 0.77 | 2.80 |
| Poverty | 0.83 | 3.66 |
| Unemployment | 0.73 | 3.02 |
| No high school diploma | 0.81 | 3.55 |
| Lack of physical activity | 0.92 | 8.82 |
| Crime | 0.37 | 1.68 |

[a]VIF: variance inflation factor.

**Table 3.** Support vector regression data set–level feature importance score.

| Features | SVR[a] feature importance |
|---|---|
| Low access to supermarket | 4.39 |
| Black | 68.20 |
| Poverty | 78.60 |
| Unemployment | 70.16 |
| No high school diploma | 73.41 |
| Lack of physical activity | 100 |
| Crime | 0 |

[a]SVR: support vector regression.

**Figure 4.** The Shapley Additive Explanations (SHAP) value plot of the feature contribution (unscaled) for the patient's neighborhood (census tract:10300). The x-axis represents the SHAP's value, and the y-axis represents the features. The lack of physical activity and poverty had the largest positive (increase) contributions to obesity prevalence in the patient's neighborhood.

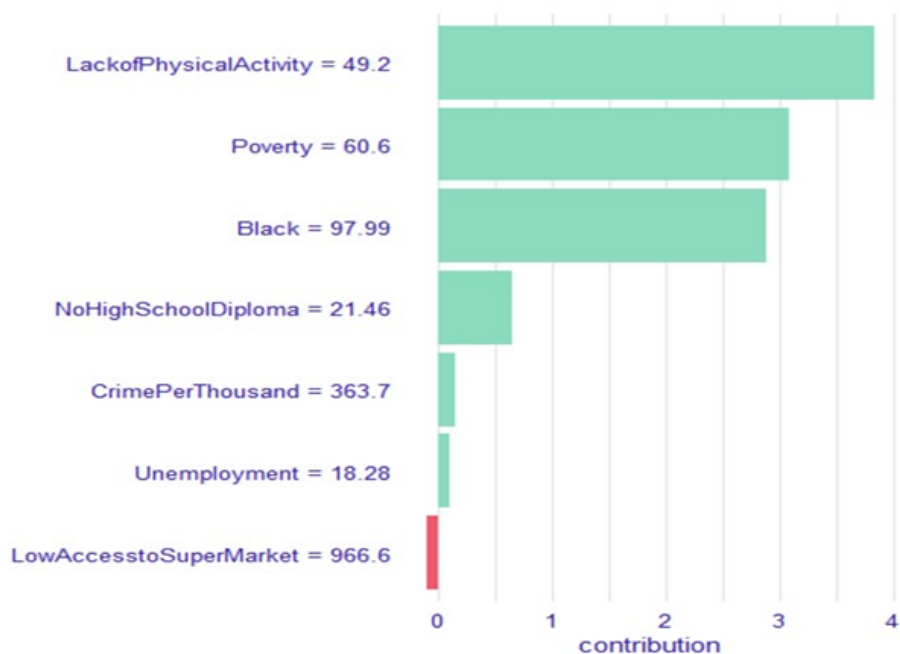

## UPHO: Dashboard Design

### Use Case Scenarios

In this section, we describe the semantics feature provided by UPHO through a proof-of-concept prototype that will display the different features of the expanded system by implementing the clinical scenarios described in the previous section.

First, the user will sign into the UPHO platform dashboard, which will determine their specific role and establish the proper access permissions. The user will make the selections from the following menu items:

- S1. Select an outcome of interest (eg, obesity prevalence, cancer,)
- S2. Select analytics aim
- S3. Select level of analysis and enter address/location (patient's address [patient-level], city, county, or state [population-level])
- S4. Select geographical level of granularity (eg, zip code, census tract)
- S5. Select SDoH domain-specific risk factors

After making these selections, the system will present on-demand explanations of risk level calculations, based on the selected level of geographic granularity.

### Scenario 1

The physician selects "obesity prevalence" as the outcome of interest (S1), and "causal pathway analysis" (S2) as the analytics aim, "patient-level" as the level of interest (entering patient's address, S3), and "census tract" as the geographical level of granularity (S4). The system provides risk-level calculations and descriptive statistics based on the census tract of the patient's address. The physician also has the option to select a particular SDoH of interest in S5, in which case the system will highlight these nodes in the graph. Finally, the user selects "Explore" to generate the results and a corresponding knowledge graph. These results are tailored to the user's interest in patient-level analysis and provide an explanative overview of the analysis results (Figure 5A). The system also allows the user to hover over pathways and nodes to explore explanative knowledge (Figure 5B, 5C) and offers a summary of recommendations and knowledge (Figure 5D).

**Figure 5.** The dashboard of the Urban Population Health Observatory displays a physician user interested in obesity prevalence in her patient's neighborhood with an overview of analysis results (A), explanations displayed when user hovers over a particular pathway (B), knowledge displayed when user hovers over a particular node (C), and summary of recommendations and knowledge (D). ACESO: adverse childhood experiences ontology; GISO: geographical information system ontology; DO: disease ontology; HIO: health indicators ontology.

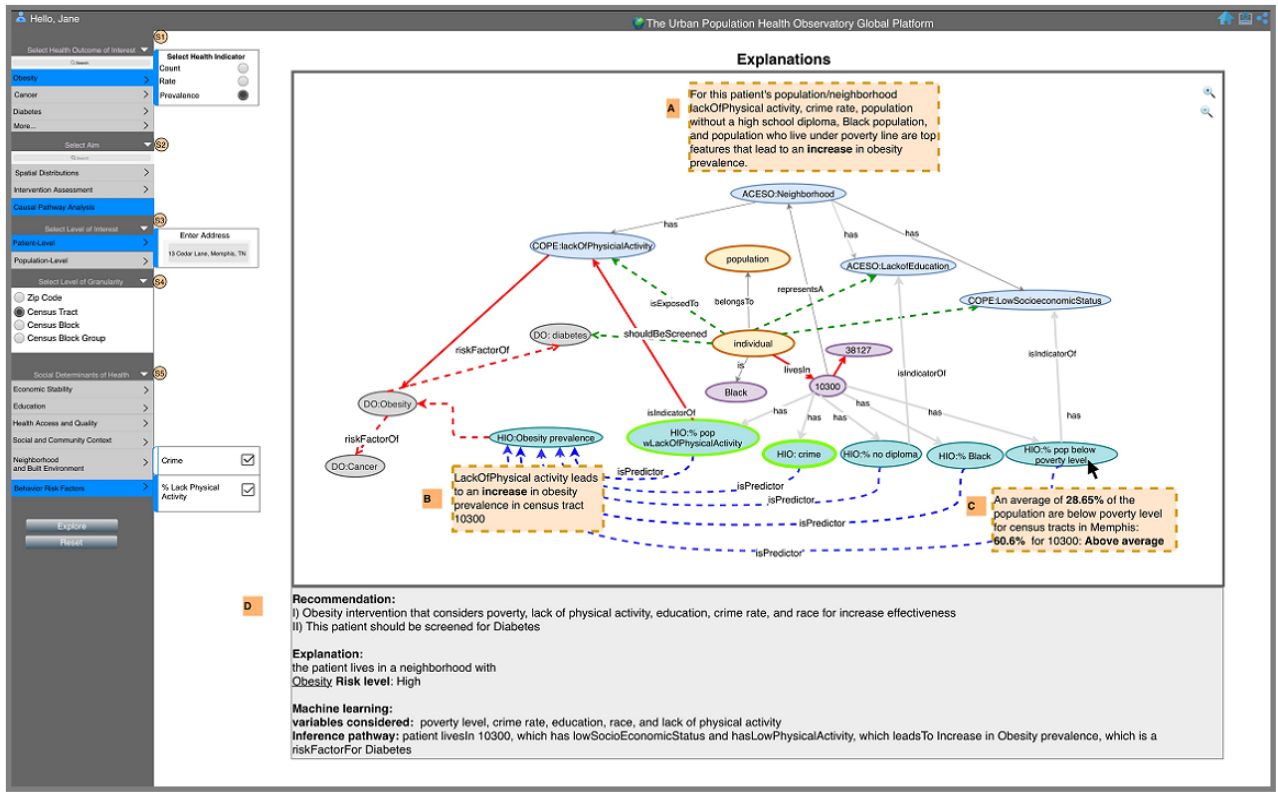

## Scenario 2

Here the researcher has access to more features. The researcher explores the causal pathway analysis aim in a population-level analysis and enters Memphis, TN, as a location of interest at the census tract–level (S1-S3), as shown in Figure 6, and the system provides risk-level calculations for the city of Memphis, TN. The researcher also has access to regression plots (Figure 6, A), which reflect the selection in S5. In section B, the system reports the results from the SVR machine model and provides explanations for each feature included in the model (Figure 6, B). In section C, the explanation pane presents a knowledge graph showing results tailored to the user's interest in population-level analysis (Figure 6, C). The researcher can also hover over pathways and nodes for knowledge (Figure 6, C, a, b, and c), like the physician in scenario 1. The system also offers the researcher a summary of recommendations and knowledge (Figure 6, C, d).

**Figure 6.** The dashboard of the Urban Population Health Observatory displays a researcher as the user interested in obesity prevalence in Memphis, TN, with univariate regression plot (A), multivariate analysis (B), and (C) which contains an overview of analysis results (a), explanations displayed when user hovers over a particular pathway (b), knowledge displayed when user hovers over a particular node (c), and summary of recommendations and knowledge (d). ACESO: adverse childhood experiences ontology; GISO: geographical information system ontology; DO: disease ontology; HIO: health indicators ontology.

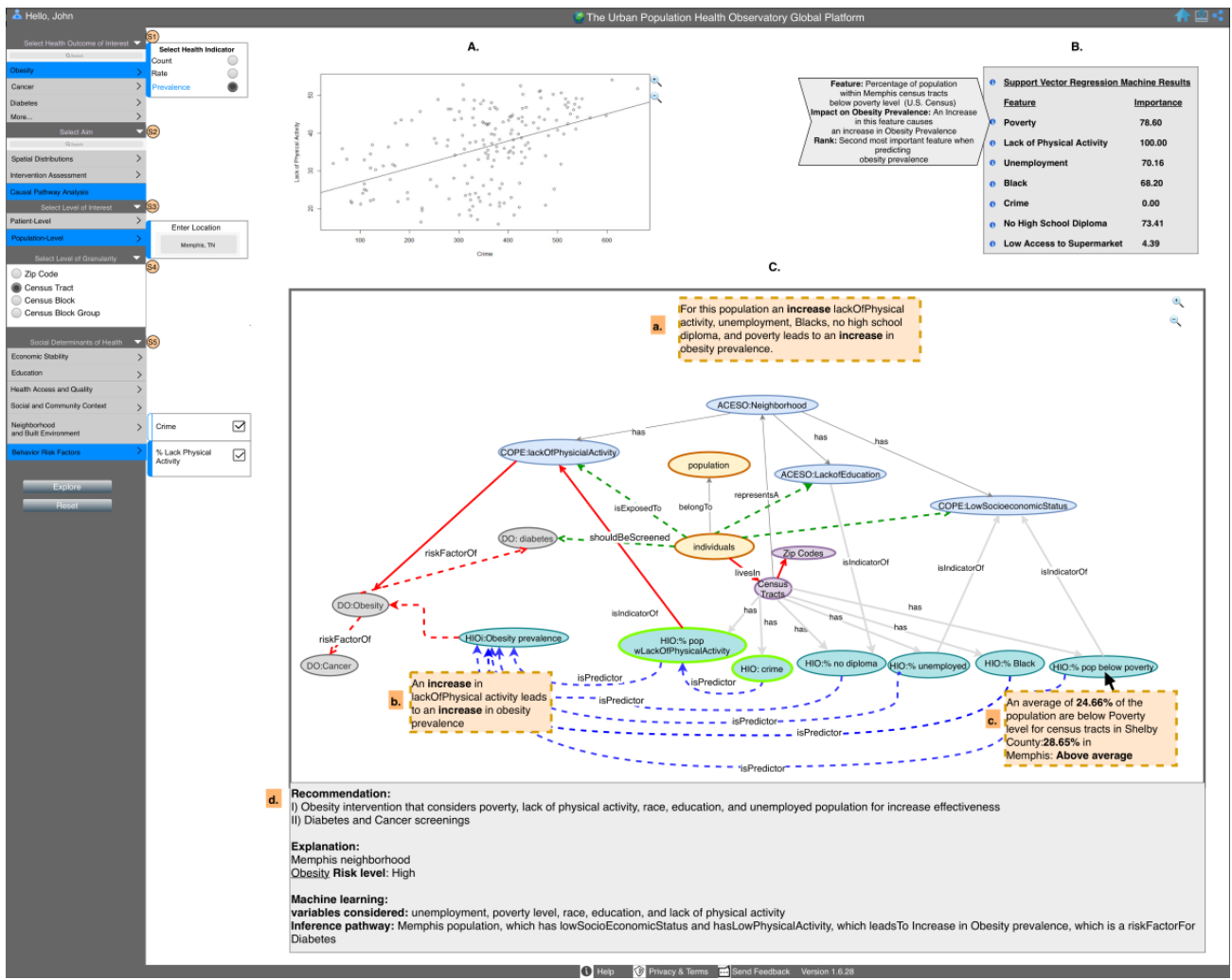

## Explanations

The graph part of the dashboard can serve as a tool for researchers and physicians to semantically explain the recommendations that we made about a specific patient or population. The current version of the graph provides 2 different visual cues, as follows.

- Tracing pathways on the graph provides visual cues. The red arrows in Figure 3 show the edges that are part of a causal pathway that leads from risk factors to negative health outcomes for the specific patient, zip code, or census tract. While this path is specific to the selected patient or population, it can be used as a generic metapath. For example, Individual *livesIn* CensusTract→*representsA* Neighborhood→*hasPhysicalCharacteristic*→*RiskFactorFor* Disease. Depending on the level of sophistication desired, the user can trace paths on the graph and click on certain nodes or edges to obtain more insights, including statistically derived evidence or semantically inferred knowledge. They can also track the sources of that knowledge including the ontologies used.

- Clicking on a node or edge on the graph displays analysis results or knowledge. The user can hover over a certain edge (eg, lackOfPhysicalActivity isPredictorOf ObesityPrevalence; Figure 5B) to obtain an explanation of the data that show lackOfPhysical activity in the patient's census tract leads to an increase in the prevalence of obesity. Similarly, the user can hover over a metric node (eg, percentage of the population below the poverty line, Figure 5C) to explain that this patient lives in a neighborhood in which nearly 61% of the population lives below the poverty line, compared to the average in their city, county, or state.

UPHO's metrics can be implemented into the backend of EHR systems (eg, Epic), and the results of those metrics can be rendered on the EHR interface in the form of risk scores on dashboards with severity indicators based on thresholds. Physicians can examine these metrics at the population level or individual patient level. UPHO can alternatively be used in a standalone approach by allowing a physician to extract more details about a single patient by providing the patient's address or a population of patients by providing their city, state, or county. The input is coded to a geographical level of granularity that can be aligned with the population-level data to gain insights into the patient's environment.

## Discussion

### Principal Findings

Previous studies provided evidence that socially disadvantaged communities are disproportionately affected by chronic diseases such as obesity [5-8], which is a risk factor for developing diabetes, heart disease, and cancer. The significance of UPHO lies in its ability to provide a multifaceted surveillance system design that serves as an apparatus for actualizing effective interventions, addressing concerns in health disparities, providing awareness to the public, and equipping health officials with a surveillance system that will improve population health decision-making and planning [24]. Using the semantics layer, the UPHO platform provides contextual knowledge by reusing several ontologies focusing on public health (eg, diseases, transportation, geography).

The incorporation of semantics provides the user with an additional layer of explainability and interpretability, which could decrease errors in intervention or treatment due to misinterpretation or misunderstanding. The semantics layer can also use ontologies to overcome the challenge of scattered data sources, thereby assisting in the achievement of interoperability, which will be used to maintain features such as data integration, XAI, and interpretability. We apply logical reasoners to extract and supply knowledge despite limited data.

Similar chronic disease surveillance systems [10-16] have offered approaches to assist in the efforts of improving chronic disease surveillance. Several published systems do not incorporate SDoH data [10-14]. One did not provide an implementation of the systematic framework [16] and several of them did not include XAI as a feature.

We followed the conceptual framework for UPHOs [17] and sought to improve the quality of disease surveillance by incorporating advances in AI and Bog Data, including interactive dashboard design, explainability, data integration, and interoperability, and incorporation of multimodal SDoH data.

Developing a multidimensional scalable surveillance system to monitor and detect trends and deliver rapid early warnings and recommendations could assist health officials, physicians, and researchers in mitigating a health crisis such as the ongoing COVID-19 pandemic [24].

### Limitations

One of the major limitations of UPHO is that it collects population data so that neighborhood or population assumptions are made for an individual in a clinical setting. For instance, individuals or patients who live in a particular population or neighborhood might not have the same characteristics as other individuals residing in the same neighborhood or population. However, our platform provides an end-to-end approach to examining the environment one resides and incorporates information that is important for the implementation of effective interventions for a given disease.

The future work will be focusing on the further development of the UPHO platform, so it can enable timely, insight-driven decisions and inform immediate or long-term health policy responses [15] to current and future public health crises.

### Conclusions

This study leveraged semantic technology and presented a proof-of-concept prototype design for our knowledge-based surveillance system, UPHO, which aims to reduce health disparities and improve population health. The expanded feature incorporates another level of interpretable knowledge needed to inform physicians, researchers, and health officials' decision-making process at the community level. Incorporating XAI helps with the explainability and interpretability of the relevant data, information, and knowledge. Users who are not equipped with domain knowledge could extract common sense knowledge from a system that incorporates XAI [35]. We as humans need a clear visualization and understanding of relationships between parameters in a system to make informed decisions. The lack of understandability and explainability in the health care and public health domains often leads to poor transparency, lack of accountability, and ultimately lower quality of care and biased health policies [39]. Thus, the incorporation of semantics and XAI can improve fairness, accountability, transparency, and trust in health care and public health.

### Acknowledgments

The funding for this study is provided by the University of Tennessee Health Science Center.

## Conflicts of Interest

None declared.

## Abbreviations

**ACE:** adverse childhood experience
**ACESO:** adverse childhood experiences ontology
**CDC:** US Centers for Control and Prevention
**COPE:** Childhood Obesity Prevention (Knowledge) Enterprise
**DO:** disease ontology
**EHR:** electronic health record
**GISO:** geographical information system ontology
**HIO:** health indicators ontology
**mHealth:** mobile health
**RMSE:** root mean square error
**SDoH:** socioenvironmental determinants of health
**SHAP:** Shapley Additive Explanations
**SVR:** support vector regression
**UPHO:** Urban Population Health Observatory
**VIF:** variance inflation factor
**WHO:** World Health Organization
**XAI:** explainable artificial intelligence

*Edited by A Mavragani; submitted 29.12.21; peer-reviewed by S Mukherjee, J Ye, D Surian; comments to author 13.04.22; revised version received 03.05.22; accepted 07.06.22; published 20.07.22*

*Please cite as:*
*Brakefield WS, Ammar N, Shaban-Nejad A*
*An Urban Population Health Observatory for Disease Causal Pathway Analysis and Decision Support: Underlying Explainable Artificial Intelligence Model*
*JMIR Form Res 2022;6(7):e36055*
*URL: https://formative.jmir.org/2022/7/e36055*
*doi: 10.2196/36055*
*PMID:*